\documentclass[10pt,twocolumn,letterpaper]{article}

\usepackage{cvpr}              

\newcommand{\TODO}[1]{\textbf{\color{red}[TODO: #1]}}
\renewcommand{\TODO}[1]{}

\usepackage{microtype}
\usepackage[accsupp]{axessibility}  

\renewcommand{\paragraph}[1]{\vspace{.5em}\noindent\textbf{#1.}}

\usepackage{soul}
\setuldepth{foobar}

\definecolor{cvprblue}{rgb}{0.21,0.49,0.74}
\usepackage[pagebackref,breaklinks,colorlinks,allcolors=cvprblue]{hyperref}
\usepackage{bm}
\usepackage{balance}
\usepackage{multirow}

\def\paperID{*****} 
\def\confName{CVPR}
\def\confYear{2026}

\title{\textsc{DiffClean}: Diffusion-based Makeup Removal for Accurate Age Estimation}

\author{Ekta Gavas\\
New York University\\
{\tt\small eg4131@nyu.edu}
\and
Sudipta Banerjee\\
University of Wyoming\\
{\tt\small sbanerj3@uwyo.edu}
\and
Chinmay Hegde\\
New York University\\
{\tt\small chinmay.h@nyu.edu}
\and
Nasir Memon\\
New York University\\
{\tt\small memon@nyu.edu}
}

\begin{document}
\maketitle
\begin{abstract}
   Accurate age verification can protect underage users from unauthorized access to online platforms and e-commerce sites that provide age-restricted services. However, accurate age estimation can be confounded by several factors, including facial makeup that can induce changes to alter perceived identity and age to fool both humans and machines. In this work, we propose \textsc{DiffClean} which erases makeup traces using a text-guided diffusion model to defend against makeup attacks. \textsc{DiffClean} improves age estimation (minor vs. adult accuracy by 5.8\%) and face verification (TMR by 5.1\% at FMR=0.01\%) compared to images with makeup. Our method is robust across digitally simulated and real-world makeup styles, and outperforms multiple baselines in terms of biometric and perceptual quality. Our codes are available at \url{https://github.com/Ektagavas/DiffClean}.

\end{abstract}

\section{Introduction} 
\textbf{Motivation.}
Our constant need to stay digitally connected has given rise to a wide range of online services, including social media platforms, chat rooms, and dating websites. Several online services rely on \textit{digital age verification}, typically age verification from selfie-styled facial scans (\textit{e.g.}, \href{https://www.yoti.com/business/facial-age-estimation/}{Yoti}, \href{https://ondato.com/age-verification/}{Ondato}, \href{https://truststamp.ai/age-estimation.html}{TrustStamp}). But there have been reports of minors bypassing age verification checks, exposing them to cyberbullying and online dangers.~\footnote{\href{https://www.dailymail.co.uk/sciencetech/article-9192397/Children-easily-bypass-age-checking-measures-social-media-sites.html}{Children bypass age-checking measures on social media sites}}\footnote{\href{https://www.scimex.org/newsfeed/dangers-for-underage-teens-bypassing-age-restrictions-on-dating-apps}{Underage teens bypassing age restrictions on dating apps.}} In such scenarios, precise age estimation is critical in protecting underage users. Recently, government agencies and commercial content providers are actively pushing for online facial age verification for protecting teenagers (US's Identifying Minors Online, UK's Online Safety Act, YouTube)~\cite{congress_2024, youtube_2025} from unauthorized access to adult-themed websites.

Age estimation is intrinsically challenging due to complex genetic, environmental, demographic, and subjective factors; this is compounded when an individual puts on makeup. Facial makeup significantly alters the perception of facial features, affecting reliable age prediction~\cite{feng2012quantifying,russell2019differential,davis2022your}. The perceived age in younger people with makeup is generally overestimated, while in older people the predicted age is underestimated \cite{voelkle2012let, watson2016you}. Makeup can be used to bypass automated age estimation~\cite{Chen2014ImpactOF}, the primary issue addressed in this work, but we also look at its impact on face recognition~\cite{ueda2010influence}. 
Intuitively, we can design a makeup-invariant age estimator and face matcher, specifically designed for minors by using ``more data''. However, this immediately raises privacy concerns: we will need data on minors wearing makeup. The next strategy might be to generate synthetic makeup images for training new models. But makeup can vary in terms of intensity, texture, color, region of application, and subjective preferences. So, synthetic data may not capture the full diversity of facial makeup.

\noindent \textbf{Our goal.} We propose to defend against makeup attacks by designing a \textit{makeup removal framework}, known as \textsc{DiffClean}. It will be deployed as a plug-in module that will filter the selfie-styled facial input image to erase makeup traces, which will primarily benefit accurate age estimation, and secondary tasks like identity verification.

\begin{figure}[t]
    \centering
    \includegraphics[width=0.98\linewidth]{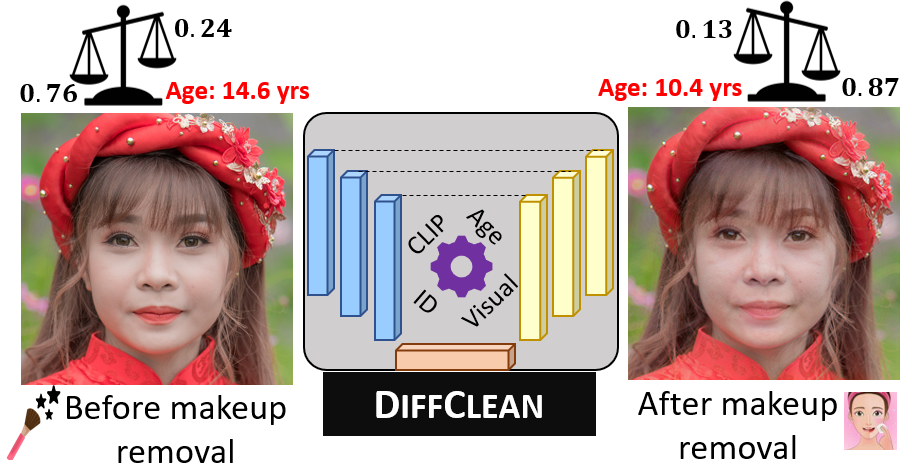}
    \caption{Overview of our method, \textsc{DiffClean}, that removes makeup on an example FFHQ image using text-guided diffusion model with a combination of CLIP loss, identity (ID) loss, Age loss and Perceptual (Visual) losses. Successful makeup removal results in shifting the softmax probabilities: $[p(\text{with makeup}), p(\text{without makeup})]$ from $[0.76, 0.24]$ to $[0.13, 0.87]$. Here, we use ViT-B/32 as CLIP-based classifier. Makeup removal reduces age overestimation from 14.6 to 10.4 years.}
    \label{fig:overview}
\end{figure}

\noindent \textbf{Our approach.} We formulate makeup removal as an image translation task, leveraging generative models known for seamless and controllable digital makeup transfer~\cite{li2018beautygan, chen2019beautyglow, jiang2020psgan, nguyen2021lipstick, yang2022elegant, shamshad2023clip2protect, sun2024diffam}. Unlike prior works that rely on attribute editing or reference-based adversarial transfer, our method \textsc{DiffClean} is a \textit{reference-free}, text-guided diffusion model that transforms a face image \textit{with} makeup to a face image \textit{without} makeup using a combination of CLIP loss (for makeup detection), perceptual loss (for fidelity retention), biometric loss (for identity restoration), and age loss (for age restoration); see Fig.~\ref{fig:overview}. Our method primarily enhances age estimation accuracy by removing makeup traces from face images when present. It can also improve facial identity verification.

\noindent \textbf{Contributions.}
Our main contributions are as follows:
\begin{enumerate}
    \item We propose \textsc{Diffclean}, a novel \textit{reference-free} text-guided diffusion-based makeup removal framework that digitally erases the traces of makeup present in face images, and improves age estimation (minor vs. adult accuracy by 5.8\%) and face verification (TMR by 5.1\% at FMR=0.01\%) compared to images with makeup, while retaining visual quality (SSIM=0.98, PSNR=35.64).
    \item We utilize text-guided CLIP loss for detecting makeup, perceptual loss for minimizing visual artifacts, and a combination of age and identity losses for restoring the facial cues affected by makeup. The combined losses ensure that our model works on both makeup and non-makeup images without producing artifacts.
    \item We empirically demonstrate that \textsc{DiffClean} works on both digitally simulated and real makeup images across variations in pose, illumination, facial accessories, and makeup styles, while exhibiting demographic fairness. Our method can potentially assist with digital age verification to protect underage users from cyber harms.
\end{enumerate}

\textbf{Responsible Usage.} We recognize that our application focus is a sensitive topic, particularly since any modeling or algorithmic errors may adversely impact a vulnerable demographic (minors/teenagers). We strongly advocate for the ethical use of \textsc{DiffClean} \textit{only} to assist with facial analytics (not for the purposes of malicious image editing).  

\section{Related Work}
\label{sec:related-work}

\hspace{0.8em}\textbf{Facial Age Prediction.} ~\cite{lanitis2004comparing} used statistical face models generated by applying Principal Component Analysis (PCA) on a set of face images for age estimation. Texture features such as Local Binary Pattern (LBP) and its variants have been used for reliable age estimation~\cite{LBP_2008, Localfeatures_2010, LBP_2012}. Biologically-inspired features have been used in~\cite{biofeatures_2009, HumansvsMC_2013}.  
Deep-learning -based approaches rely on convolutional neural networks (CNNs) and transformers for improved robustness. 
A multi-stream CNN was used to learn high-dimensional structured information from face patches in~\cite{yi2014age}. Deep CNN was used for age estimation in unconstrained settings in~\cite{ranjan2015unconstrained, wang2015deeply, levi2015age}. In~\cite{niu2016ordinal}, each age was treated as a rank to apply ordinal regression problem to age estimation using a deep CNN. A multi-task learning approach for heterogeneous attribute estimation was adopted in~\cite{han2017heterogeneous}. Deep Expectation (DEX)~\cite{rothe2018deep} formulated age estimation by converting it into a classification-regression problem. SSRNet~\cite{yang2018ssr} adopted a compact soft stagewise regression model to perform multi-class classification. Auxiliary facial cues were used to boost reliable age estimation in~\cite{liu2020facial,zeng2020soft}. Deep random forests were used for age estimation in~\cite{DeepForests_2021}. FP-age~\cite{lin2022fp} used a face parsing attention module to incorporate facial semantics into age estimation. 
Swin Transformer with attention-based convolution (ABC) for facial age estimation was proposed in~\cite{Shi_2023}. In~\cite{kuprashevich2023MiVOLO}, a multi-input transformer-based model, MiVOLO, predicted age and gender using facial and full body information.


\textbf{Age Estimation in Minors.} Limited works have focused on improving age estimation performance in minors. In~\cite{antipov2016apparent}, a finetuned VGG-based `children' network was used for apparent age estimation of children aged from 0 to 12 years. \cite{anda2020deepuage} trained ResNet50 on the underage facial dataset `VisAGe' to improve the age estimation for minors. The need for fast, accurate, and robust age estimation and verification has gained traction recently. NIST Face Analysis Technology Evaluation (FATE) Age Estimation \& Verification reports that subjects aged between 8 and 12 years tend to be overestimated to be 13 to 16 years old with false positives as high as $0.76$ \footnote{\href{https://pages.nist.gov/frvt/html/frvt_age_estimation.html}{NIST FATE Report}}; higher errors are reported for females than for males, specifically, in 17-year-olds~\cite{NIST_FATE_2024}. The performance of facial age estimation algorithms can be affected by several factors like gender, ethnicity, image quality, facial expression, and makeup~\cite{anda2020assessing, clapes2018apparent}.

\textbf{Impact of Cosmetic Makeup on Facial Analytics.}~\cite{ueda2010influence} demonstrated the impact of facial makeup on automatic face recognition.~\cite{feng2012quantifying} observed that makeup applied to different facial regions had varying effects on age prediction quantified as Young Index (YI). \cite{eckert2013facial} observed that the Identification Rate (IDR) decreased with increasing amount of makeup. ~\cite{chen2013automatic,guo2013face,hu2013makeup} focused on making face recognition more robust to makeup. Authors in~\cite{Chen2014ImpactOF} observed that facial cosmetics (both real makeup and synthetically retouched) affect automated age estimation, causing a mean absolute error of up to 5.84 years. They further indicated that lip makeup had the least impact on automated age estimation, followed by eye makeup, followed by full face makeup (eye + lip + foundation). 
Makeup produces differential effects on apparent (manual) age estimation. 20-year-old women appeared older when wearing full face makeup, while 40- and 50-year-old women appeared younger when wearing makeup~\cite{Dayan2015QuantifyingTI, clapes2018apparent, russell2019differential, davis2022your}. Few works discuss the influence of makeup on facial age estimation in minors~\cite{anda2020assessing, anda2020deepuage}. 

\textbf{Facial Makeup Transfer and Removal.}
Approaches for makeup transfer relied on GANs for style-transfer networks~\cite{li2018beautygan, chen2019beautyglow, jiang2020psgan, nguyen2021lipstick, yang2022elegant, sun2024content}. Later, diffusion models were used~\cite{zhang2024stable, sun2024shmt,ruan2025mad}.
Makeup removal, on the other hand, poses a more challenging task where the goal is to reveal the face under makeup without adversely affecting the biometric and perceptual features. Anti-makeup~\cite{li2018anti} integrated two adversarial networks, both at pixel and feature levels, to generate identity-preserving synthetic non-makeup faces. LADN~\cite{ladn_2019} used local disentanglement with adversarial network for makeup transfer and removal.  ~\cite{zhang2021makeup} replaced convolution layers in GAN with resnet blocks to design makeup removal network. PSGAN++~\cite{PSGAN++_2021} used pose and expression robust spatial-aware GAN for detail-preserving makeup transfer and removal.  CLIP2Protect~\cite{shamshad2023clip2protect} and DiffAM~\cite{sun2024diffam} use adversarial makeup transfer for privacy with capability to generate `no makeup' faces. \cite{ruan2025mad} used domain translation for a wide array of makeup tasks, including makeup transfer and removal as a one-for-all diffusion-based framework, but suffers from lack of generalizability. DeBeauty~\cite{DeBeauty_2025} is the \textit{only} reference-free approach to the best of our knowledge that uses adversarial de-makeup flow and relocation deformation flow to remove makeup. 
\section{Proposed Method}
\label{sec:3}

\begin{figure}[t]
    \centering
    \includegraphics[width=0.75\columnwidth]{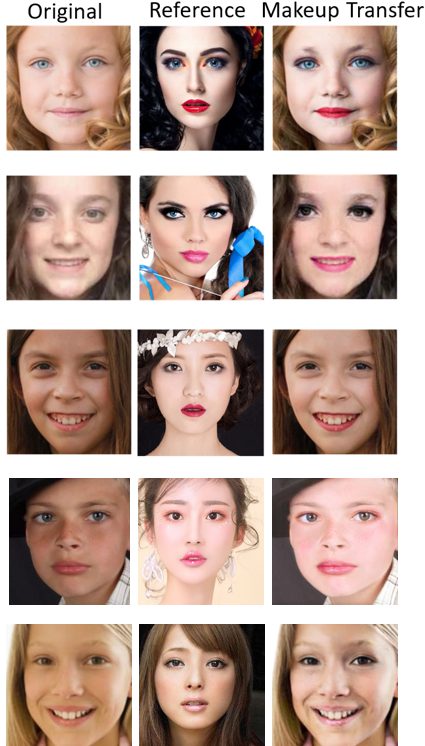}
    \caption{Synthetic makeup transfer results produced using EleGANt~\cite{yang2022elegant} on example images from UTKFace. Original image (first column), Makeup style reference (second column), and Makeup transferred image (third column).}
    \label{fig:sub-utk}
\end{figure}

Our proposed method specifically targets mitigating the impact of makeup on automated facial age estimation and, additionally, facial identity verification. Defending against makeup attacks can prevent minors from bypassing age verification intended for protection against cyber abuse. One strategy may involve collecting data of minors wearing makeup for improving the robustness of age estimators; this poses strong privacy concerns. So, we propose a generative model \textsc{Diffclean}, which is designed to mitigate age variations arising due to makeup. There are two challenges in designing this network: (1) we need a robust \textit{makeup style generator} to synthesize data needed for training the makeup removal model and thus circumvent data collection; (2) we need a robust \textit{age estimator} to serve as an auxiliary classifier to guide \textsc{DiffClean} to restore age-specific cues. For the \textit{makeup style generator}, we use EleGANt~\cite{yang2022elegant} for locally-editable makeup transfer; see Fig.~\ref{fig:sub-utk}. For the \textit{age estimator}, we use SSRNet~\cite{yang2018ssr} fine-tuned on UTKFace~\cite{UTKFace} dataset that has images from different age groups. 

\textsc{DiffClean} uses the ablated diffusion model (ADM) with guided diffusion from~\cite{dhariwal2021diffusion} with improved U-Net architecture containing increased attention heads at multiple resolutions, BigGAN residual stacks for upsampling and downsampling, deeper networks, and rescaled residual connections; refer to ~\cite{dhariwal2021diffusion} for details. We used text prompts \textit{face with makeup} and \textit{face without makeup} for text-guided CLIP directional loss following~\cite{sun2024diffam} that detects if makeup is present or not. Makeup removal relies on the combined effect of CLIP-based makeup detection, perceptual utility retention, and restoration of identity and age-specific cues.   




\noindent \textbf{Methodology.}
We describe the terminology and notation here. Original/Clean image (without makeup), $\textbf{I}_o$, Makeup image (with real/synthetic makeup), $\textbf{I}_m$, and Makeup removed image, $\textbf{I}_g$. Consider $\mathcal{M}$ as a real or synthetic makeup application operator, while $\mathcal{M}^{-1}$ represents the makeup removal operator. Thus, $\textbf{I}_o \xrightarrow[]{\mathcal{M}} \textbf{I}_m$, while $\textbf{I}_m \xrightarrow[]{\mathcal{M}^{-1}} \textbf{I}_g$. The goal is to model $\mathcal{M}^{-1}$ so that we can recover the clean image, $\textbf{I}_g \approx \textbf{I}_o$. Our proposed framework \textsc{DiffClean} approximates the role of $\mathcal{M}^{-1}$. 

\noindent \textbf{Architecture:} The makeup removal framework ($\mathcal{M}^{-1}$) used in this work is a generative model which is a pre-trained ADM~\cite{dhariwal2021diffusion} that adds noise to the input image (with makeup), $\textbf{I}_m$, to generate the noisy latent image during forward diffusion. The deterministic reverse diffusion through DDIM produces the makeup-removed output, $\textbf{I}_g$ supervised by CLIP, biometric identity, age, and perceptual losses. 

\noindent \textbf{Loss functions.} We present the loss functions used for supervising the age estimator and the diffusion model. 

\noindent \textbf{Age Estimator:} SSRNet is fine-tuned using $\mathcal{L}_{WSL}$, a weighted adaptation of self-adjusting smoothed-$\mathcal{L}_1$ loss. The adaptive weight ensures stricter penalty for misclassification in vulnerable age groups [10-29] years as follows:
\begin{align}
    \mathcal{L}_{WSL} = \begin{cases}
			3.0 * \mathcal{L}_{SL}, \text{if age $\{10-29\}$ yrs.}\\
            1.0 * \mathcal{L}_{SL}, \text{otherwise.}
            \end{cases}
\end{align}


Refer to~\cite{fu2019retinamask} for details about $\mathcal{L}_{SL}$. Note we ablated with different losses such as $\mathcal{L}_1$, $\mathcal{L}_2$ and Huber loss functions, but empirically observed $\mathcal{L}_{WSL}$ loss to perform the best.

\noindent \textbf{\textsc{DiffClean}:} During \textit{training}, we used four losses to fine-tune the diffusion model.

\textit{CLIP-based Makeup Loss:} 
The transition from makeup domain to non-makeup domain is guided via textual prompts `face without makeup' and `face with makeup' in CLIP space using ViT-B/16 encoder following~\cite{sun2024diffam}. We define the loss as one minus the cosine similarity between the $\mathcal{L}_2$-normalized image and text embeddings as follows: 
\begin{equation}
    \mathcal{L}_{clip}  = 1 - \frac{\Delta \textbf{I} .\Delta \textbf{T}}{\left \lVert \Delta \textbf{I} \right \rVert \left \lVert \Delta \textbf{T} \right \rVert}
    \label{eqn-clip_loss}
\end{equation}
\noindent
where \(\Delta \textbf{I} = \mathcal{E}_\textbf{I}(\textbf{I}_g) - \mathcal{E}_\textbf{I}(\textbf{I}_m)\) and \(\Delta \textbf{T} = \mathcal{E}_\textbf{T}(t_{\text{face w/o makeup}}) - \mathcal{E}_\textbf{T}(t_{\text{face w/ makeup}})\). Here, \(t_{\text{face w/o makeup}}\) is textual prompt after makeup removal and \(t_{\text{face w/ makeup}}\) is textual prompt before makeup removal. Here, \(\mathcal{E}_\textbf{I}\) and \(\mathcal{E}_\textbf{T}\) are CLIP image and text encoders, respectively.

\textit{Biometric identity Loss:}
Our goal involves identity restoration of the makeup-removed image with respect to the original image (without makeup). During training with synthetic makeup pairs, we have access to ($\textbf{I}_o$, $\textbf{I}_m$) to generate $\textbf{I}_g$. So, we compute weighted pairwise biometric identity losses as $\mathcal{L}_{og}$ and $\mathcal{L}_{mg}$. This is different than ~\cite{sun2024diffam} that uses a single identity loss between input and output.
The final identity loss is as follows:
\begin{equation}
    \mathcal{L}_{id} = \lambda_1 \mathcal{L}_{og} + \lambda_2 \mathcal{L}_{mg}
\end{equation}
where $\mathcal{L}_{\cdot, \cdot}$ represents the cosine distance between the extracted features using the ResNet ArcFace from the InsightFace library, with weights \(\lambda_1 =0.75\) and \(\lambda_2=0.25\). 

\textit{Perceptual Loss:}
To preserve the perceptual features, we use the perceptual loss $\mathcal{L}_{pips}$ with VGG \cite{simonyan2014very} network and $\mathcal{L}_1$ loss between generated makeup-removed image \(\textbf{I}_g\) and input image with makeup \(\textbf{I}_m\).  

\textit{Age Loss:}
The age loss is computed as the smoothed-$\mathcal{L}_1$ loss (introduced in RetinaMask~\cite{fu2019retinamask}) between the the ground truth age, \(a_i\) of the original image, $\textbf{I}_o$, and the predicted age, \(\hat{a_i}\) of the makeup removed image, $\textbf{I}_g$. We used the fine-tuned SSRNet model to predict \(\hat{a_i}\). 
We introduce the SSRNet age loss as follows:
\begin{equation}
    \mathcal{L}_{\text{SSRNet-age}} = \frac{1}{n} \sum_{i=1}^{n} \mathcal{L}_{SL}(a_i,\hat{a}_i )
\end{equation}

We further tested with CLIP-based age loss in lieu of SSRNet, where we probed the CLIP model with the generated image, $\textbf{I}_g$, using the prompt $\textbf{t}_{age} = \text{\textit{face of }} \{a_i\}\text{\textit{-year old}}$, where \(a_i\) indicates the ground truth age label as follows:
\begin{equation}
\mathcal{L}_{\text{CLIP-age}} = 1- \frac{\textbf{I}_g.\textbf{t}_{age}}{\left \lVert \textbf{I}_g \right \rVert \left \lVert  \textbf{t}_{age} \right \rVert}
\end{equation}

\textbf{Total loss for makeup removal:}
Combining all the above losses, the final loss for \textsc{DiffClean} is as follows:
\begin{equation}
    \begin{split}
    \mathcal{L}_{total} &= \lambda_{clip}\mathcal{L}_{clip} + \lambda_{id}\mathcal{L}_{id} + \lambda_{lpips}\mathcal{L}_{lpips} + \lambda_{L1}\mathcal{L}_{L1} \\ &+ \lambda_{age}\mathcal{L}_{age} 
    \end{split}
    \label{Eqn:totalloss}
\end{equation}
where loss weights are tuned as \(\lambda_{clip}=5.0\), \(\lambda_{id}=1.0\), \(\lambda_{lpips}=5.0\), \(\lambda_{L1}=2.0\), and \(\lambda_{age}=0.5\) for SSRNet age loss and \(\lambda_{age}=5.0\) for CLIP age loss. Hyperparameters were selected using grid search; see sensitivity analysis in Sec.~\ref{subsec:AddAnalysis}. Thus, we design two models: \textsc{DiffClean} with SSRNet age loss ($\mathcal{L}_{age} = \mathcal{L}_{\text{SSRNet-age}}$) and \textsc{DiffClean} with CLIP age loss ($\mathcal{L}_{age} = \mathcal{L}_{\text{CLIP-age}}$).

During \textit{testing}, our model accepts face images captured under unconstrained conditions, either without makeup or with real or synthetic makeup as input. 

\begin{figure*}[ht]
    \centering
    \includegraphics[width=0.85\textwidth]{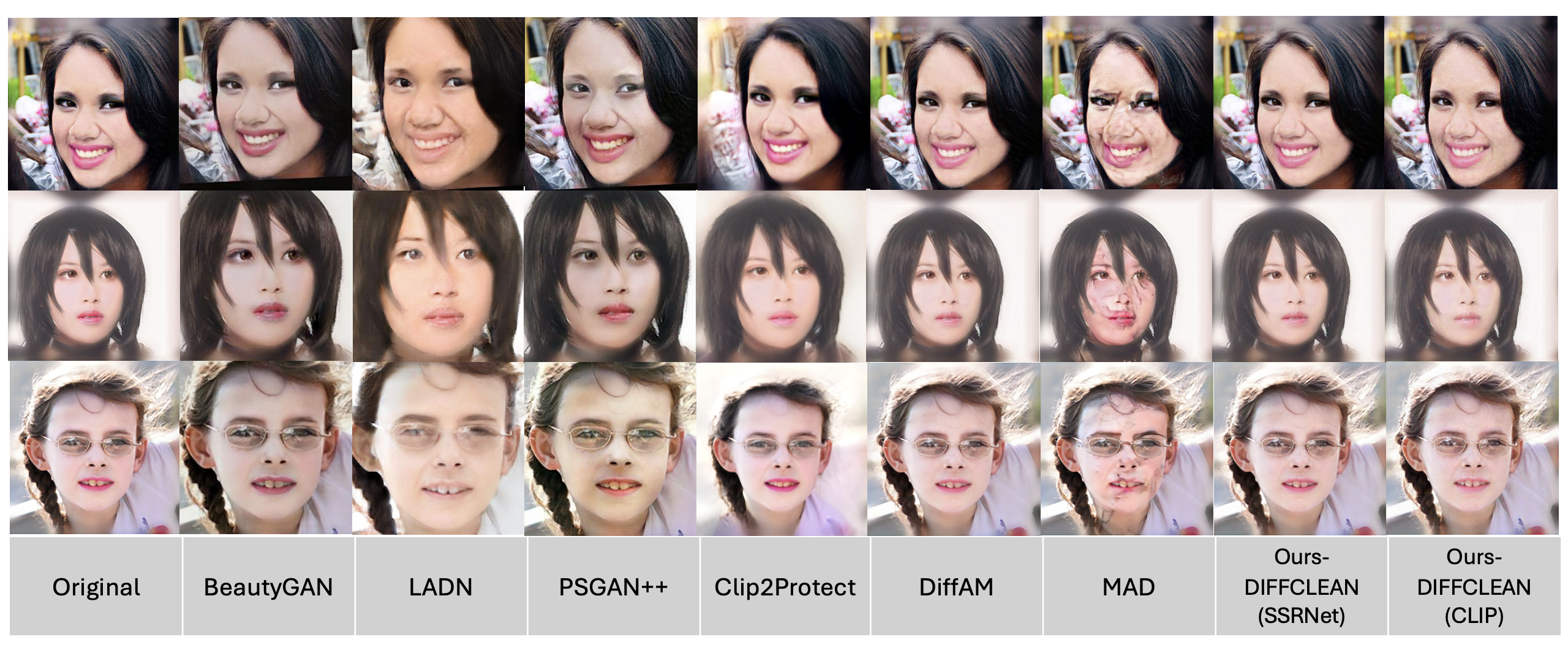}
    \caption{Comparison of makeup removal results generated by six baselines and our proposed \textsc{DiffClean} (last two columns) on three example images from FFHQ dataset. GAN-based baselines (BeautyGAN, LADN, PSGAN++) introduce visual artifacts, while CLIP2Protect alters hair color and style, DiffAM does not effectively remove makeup, and MAD produces distortions on unseen data.}
    \label{fig:qual}
\end{figure*}

\section{Experiments and Analysis}
\label{sec:Expts}
\subsection{Datasets} 
\label{subsec:Data}

 \textbf{Training/Fine-tuning:} We used the UTKFace~\cite{UTKFace} dataset to fine-tune SSRNet for age estimation with a training set of 15,364 images and a testing set of 6,701 images. The original set of 20K images was split into training and testing splits (70:30) after filtering out images with ages $\geq$70 yrs. We adopted a two-step approach for fine-tuning \textsc{DiffClean} for makeup removal. First, we used 300 images (200 training, 100 validation) from MT dataset~\cite{li2018beautygan} following the guidelines in~\cite{sun2024diffam}. We further refined the model by additionally fine-tuning it on 300 images (200 training, 100 validation) from the UTKFace dataset after applying makeup using EleGANt~\cite{yang2022elegant} with the combined loss. Thus, our training set $D_{tr}$ consisted of a total of 600 images.

\noindent \textbf{Evaluation:} We tested our proposed \textsc{DiffClean} on both \textit{synthetic} makeup (on FFHQ) and \textit{real-world} makeup (LADN, BeautyFace, Makeup-Wild) images.

\noindent \textit{\underline{Synthetic makeup data generation:}} We sampled 2,556 images from the Flickr-Faces-HQ Dataset (FFHQ)~\cite{karras2019style} by sampling 284 images from each of the 9 age groups (excluding $\geq 70$ yrs.)using the age bins provided by FFHQ-Aging~\cite{FFHQ_Aging, FFHQ_JSON}.
Next, we applied EleGANt~\cite{yang2022elegant}, which needs reference images for makeup transfer. We used five reference images~\cite{li2018beautygan} to simulate varied makeup styles, representing lip, face and eye makeup; see Fig.~\ref{fig:sub-utk}. We note that there can be other makeup styles, but the representative reference styles selected for data creation mimic real-world samples. 


\noindent \textit{\underline{Real-world makeup data curation:}} We used 3,000 images from the BeautyFace~\cite{Beautyface} dataset, 355 images from the LADN dataset~\cite{ladn_2019}, and 384 images from the Makeup-Wild dataset~\cite{jiang2020psgan} for testing generalizability of our method across variations in pose, illumination, accessories and occlusions.


\subsection{Metrics} 
We evaluated \textsc{DiffClean} for (1) \textbf{age restoration} using Mean Absolute Error (MAE) between ground truth (if available) and predicted age values (\textit{lower is better}); Age group accuracy to assess if the predicted age is correctly grouped into age bins (\textit{higher is better}); Minor/Adult Accuracy to evaluate if the predicted age $\geq 18$ to classify as adult, otherwise, minor (\textit{higher is better}). We evaluated \textsc{DiffClean} for (2) \textbf{identity verification} by reporting the True Match Rate (TMR) at a False Match Rate (FMR) of 0.01\% and illustrated the ROC curves. Finally, we computed (3) \textbf{image quality} of generated images using SSIM and PSNR.

\subsection{Implementation Details}

The SSRNet model was fine-tuned with a batch size of 50 with an Adam optimizer with weight decay of 1e-4, a cosine annealing scheduler, and a learning rate set to 1e-3 for 200 epochs with early stopping. The input image size is 64$\times$64.
The makeup removal module takes the makeup image, $\textbf{I}_m$, of size 256$\times$256 as input, and uses 40 DDIM inversion steps with 6 DDIM sampling steps (total time steps=80), and noise scale for latent image generation is controlled by cosine beta scheduler~\cite{nichol2021improved}. We used the Adam optimizer with a learning rate of 4e-3. The model is fine-tuned for 5 epochs. All experiments were carried out on a single NVIDIA A100 GPU. Hyperparameters in Eqn.~\ref{Eqn:totalloss} were obtained using grid search on the validation set (see sensitivity analysis in Sec.~\ref{subsec:AddAnalysis}). For testing, we use MiVOLO~\cite{kuprashevich2023MiVOLO} for age estimation (training uses SSRNet) and MobileFace \cite{chen2018mobilefacenets} and FaceNet \cite{schroff2015facenet} for identity verification ( training uses ArcFace IRSE50 \cite{hu2018squeeze}).

\subsection{Baselines}
We compared \textsc{DiffClean} with 6 baselines (4 GAN-based and 2 diffusion-based models): BeautyGAN~\cite{li2018beautygan}, LADN~\cite{ladn_2019}, PSGAN++~\cite{PSGAN++_2021} CLIP2Protect~\cite{shamshad2023clip2protect}, DiffAM~\cite{sun2024diffam}, and MAD~\cite{ruan2025mad}. 
We randomly selected six non-makeup images as references for baselines following \cite{DeBeauty_2025}.


\begin{figure*}[h]
     \centering
         \includegraphics[width=0.7\textwidth]{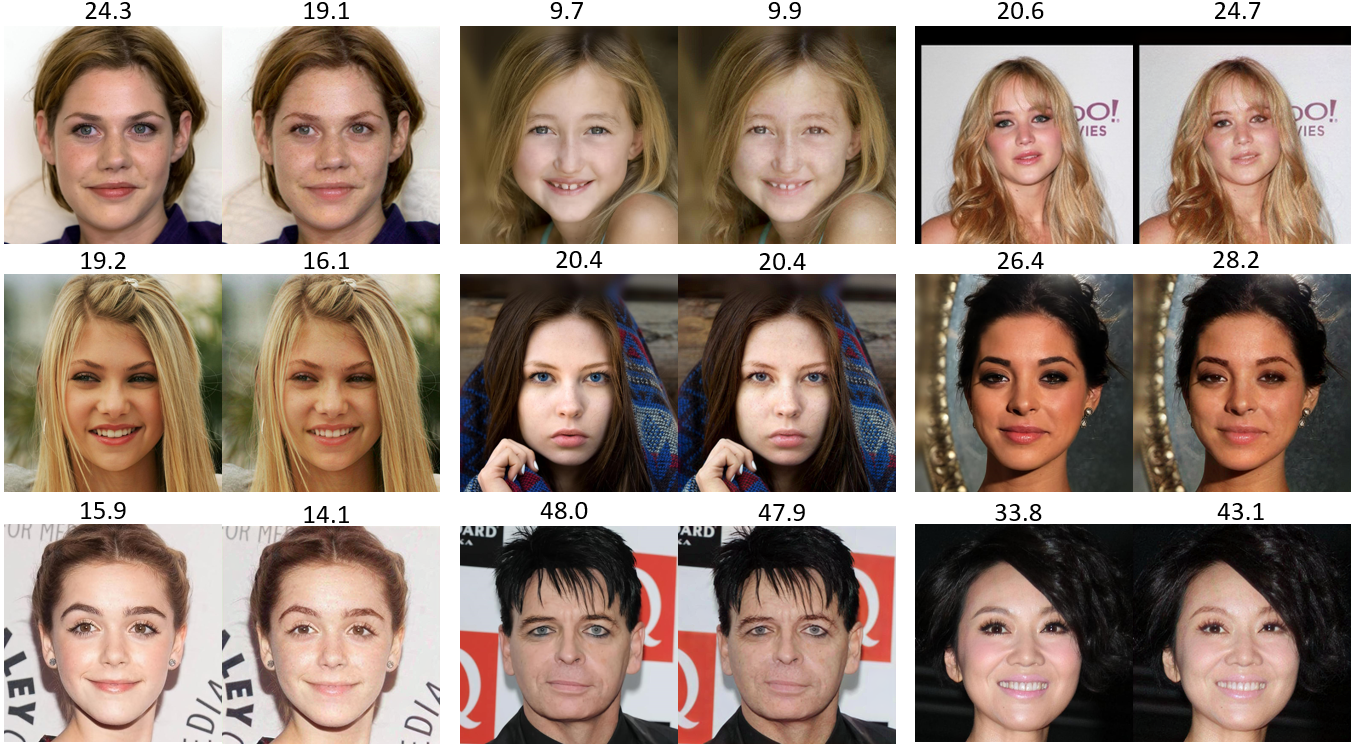} 
         \caption{Examples of differential effects of age estimation before and after makeup removal. \bm{$1^{st}$} \textbf{column:} \textsc{DiffClean} causes \textit{lower} predicted ages than with makeup, thus reducing overestimation errors; \bm{$2^{nd}$} \textbf{column:} \textsc{DiffClean} \textit{preserves} original age when there is minimal or no makeup; \bm{$3^{rd}$} \textbf{column:} \textsc{DiffClean} causes \textit{higher} predicted ages than with makeup, thus reducing underestimation errors. } 
         \label{fig:agevar}
\end{figure*}

\section{Results and Discussion}
\begin{figure}[t]
     \centering
     \begin{subfigure}[b]{0.4\textwidth}
         \centering
         \includegraphics[width=\textwidth]{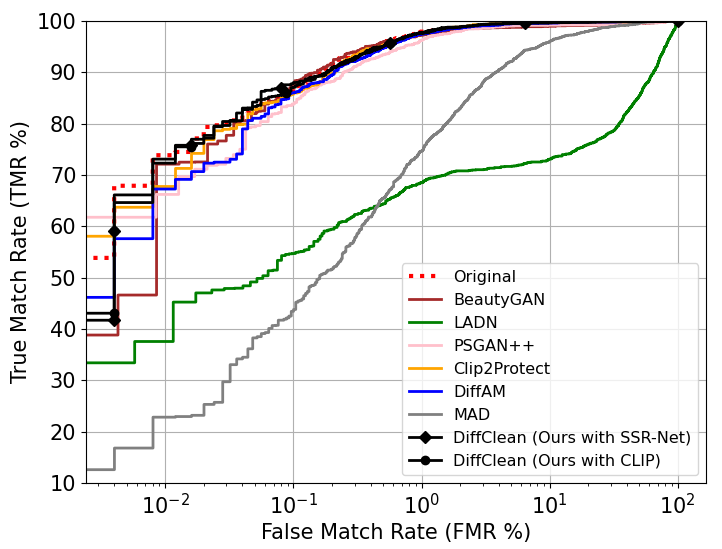} 
         \caption*{\textbf{FaceNet}}
     \end{subfigure} 
     \begin{subfigure}[b]{0.4\textwidth}
         \centering
         \includegraphics[width=\textwidth]{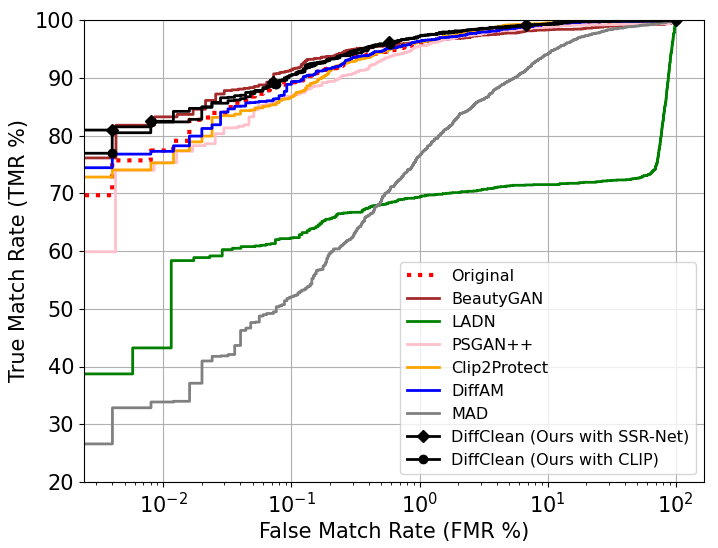}
         \caption*{\textbf{MobileFace}}
     \end{subfigure}  
    \begin{subfigure}[b]{0.48\textwidth}
    \centering
    \begin{tabular}{l||ll} \hline
     \multirow{2}{*}{\textbf{Methods}} & \multicolumn{2}{c}{\textbf{FR model}}         \\ \cline{2-3}
        & FaceNet    & MobileFace  \\ \hline \hline
\textcolor{blue}{Original}   &  74.5       &    79.1   \\
BeautyGAN   &  72.1 ($\downarrow$)        &   \underline{83.2} ($\uparrow$)   \\
LADN   &  45.2 ($\downarrow$)        &   58.3 ($\downarrow$)   \\
PSGAN++   &  66.2 ($\downarrow$)        &   75.3 ($\downarrow$)   \\
CLIP2Protect   &  67.8 ($\downarrow$)        &   75.3 ($\downarrow$)   \\
DiffAM  & 69.1 ($\downarrow$) &  78.2 ($\downarrow$)   \\
MAD  & 22.8 ($\downarrow$) &  33.9 ($\downarrow$)   \\ \hline
Ours-\textsc{DiffClean} (SSRNet)                 &   \textbf{75.8} ($\uparrow$)      & \textbf{84.2}   ($\uparrow$)              \\
Ours-\textsc{DiffClean} (CLIP)                   & \underline{75.5} ($\uparrow$)        & {82.4}  ($\uparrow$)                         \\ \hline 
\end{tabular}
    \end{subfigure}
  \caption{(Top): ROC curve with FaceNet. (Middle): ROC curve with MobileFace. (Bottom): Biometric matching in terms of TMR (\%) @FMR = 0.01\% (\textit{higher is better}) with FaceNet and MobileFace matchers on {FFHQ}. \textbf{Bolded} values indicate best results while \underline{underlined} values indicate second-best results.}
    \label{fig:biom}
\end{figure}

We present qualitative results for subjective evaluation (Figs.~\ref{fig:qual},~\ref{fig:agevar},~\ref{fig:realworld}, and~\ref{fig:covariates}). We compare with the baselines in terms of age estimation (Tables~\ref{tab:ageest1} and \ref{tab:realstats}), identity verification (Fig.~\ref{fig:biom}) and visual quality (Table~\ref{tab:visqual}).


\subsection{Age estimation}
We report the results of the predicted age estimated by MiVOLO after makeup removal on FFHQ dataset in Table~\ref{tab:ageest1}. We compute the \textcolor{blue}{Original} performance by comparing the clean FFHQ face images (without makeup) with images simulated using EleGANt (with makeup), and compute the minor vs adult accuracy (\%), age group accuracy (\%), and mean absolute error (MAE). In Table~\ref{tab:ageest1}, our method \textsc{DiffClean} outperforms all baselines and can successfully restore age cues disrupted due to makeup by lowering the MAE (by 0.85) and improving the minor/adult prediction accuracy (by 5.1\%). Our objective is focused on minor age groups and the results indicate that we achieve a superior minor vs. adult age classification of 88.6\%. 

Makeup exhibits a differential effect on facial age. It may increase or decrease apparent age and, in some cases, may not affect perceived age. \textbf{Our method does not arbitrarily reduce the predicted age in all cases}. Instead, it strategically erases makeup traces to restore the original age-specific cues. Fig.~\ref{fig:agevar} depicts the differential age restoration to minimize overestimation and underestimation errors while retaining the original age in presence of light to no makeup.

\begin{table*}[ht]
\centering
\caption{Results of age estimation on {FFHQ} in terms of minor vs adult accuracy (\%) $\uparrow$, age group accuracy (\%) $\uparrow$ and mean absolute error (MAE) $\downarrow$. \textbf{Bolded} values indicate best results while \underline{underlined} values indicate second-best results.}
\label{tab:ageest1}
\resizebox{\textwidth}{!}{%
\begin{tabular}{l||lllllll|ll}
\hline
\textbf{Metrics} &
  \multicolumn{1}{l}{\textcolor{blue}{Original}} &
  \multicolumn{1}{l}{\textbf{BeautyGAN}} &
  \multicolumn{1}{l}{\textbf{LADN}} &
  \multicolumn{1}{l}{\textbf{PSGAN++}} &
  \multicolumn{1}{l}{\textbf{Clip2Protect}} &
  \multicolumn{1}{l}{\textbf{DiffAM}} &
  \multicolumn{1}{l}{\textbf{MAD}} &
  \multicolumn{1}{|l}{\textbf{\begin{tabular}[c]{@{}l@{}}Ours-\textsc{DiffClean} \\ (SSRNet)\end{tabular}}} &
  \multicolumn{1}{l}{\textbf{\begin{tabular}[c]{@{}l@{}}Ours-\textsc{DiffClean} \\ (CLIP)\end{tabular}}} \\ \hline \hline
\textbf{Minor/Adult Acc (\%)} &  82.8 & 86.12 & 84.34 & 82.7  & 83.8 & 85.5 & 72.53 & \underline{87.8} & \textbf{88.6} \\
\textbf{Age group Acc (\%)}   & 34.4 & 35.57 & 30.9  & 32.54 & 33.3 & 34.5 & 28.03 & \underline{36.8} & \textbf{37.0}   \\
\textbf{MAE}                 & 6.56 & 6.09  & 6.98  & 6.95  & 6.54 & 6.32 & 9.51  & \underline{5.76} & \textbf{5.71} \\ \hline
\end{tabular}%
}
\end{table*}

\subsection{Identity verification}
We present the results of biometric identity verification in Fig.~\ref{fig:biom} using FaceNet and MobileFace face recognition (FR) models at FMR=0.01\%. We compute the \textcolor{blue}{Original} performance by comparing the clean FFHQ face images (without makeup) with images simulated using EleGANt (with makeup). Baseline performances are computed by comparing the original FFHQ face images with outputs of respective baselines (makeup-removed images). We observe that makeup removal with our approach improves the True Match Rate (TMR) by 1.3\% with FaceNet and 5.1\% with MobileFace compared to `Original' performance. MAD does not perform well on unseen data (trained on MT; tested on FFHQ). This manifests in the form of visual artifacts that lower overall biometric performance.

\begin{figure}[t]
     \centering
    \includegraphics[width=0.45\textwidth]{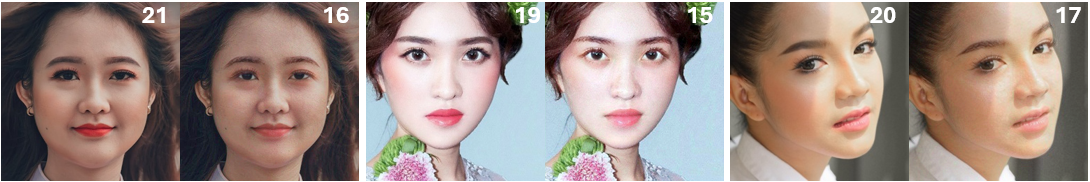}    
         \caption{Results of \textsc{DiffClean} on real-world makeup images from LADN~\cite{ladn_2019} (left), BeautyFace~\cite{Beautyface} (center), and Makeup- Wild~\cite{jiang2020psgan} (right) datasets.}  
         \label{fig:realworld}
\end{figure}

\begin{figure}[t]
\centering
\includegraphics[width=\columnwidth]{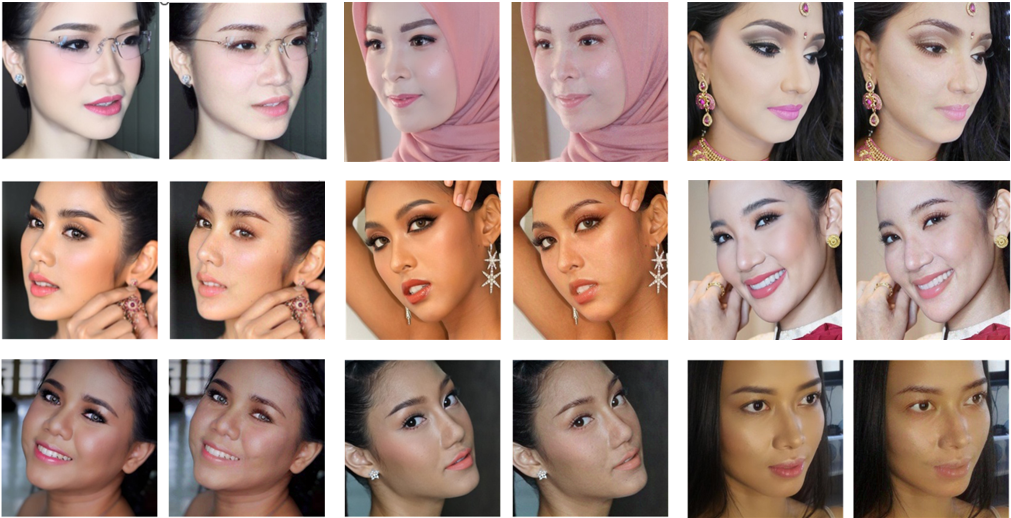}
    \caption{Examples from Makeup-Wild dataset with variations in pose, illumination, background and facial accessories. \textsc{DiffClean} preserves details such as fingers, jewelry, and expressions.}
    \label{fig:covariates}
\end{figure}

\subsection{Image quality evaluation}
We compared the images before and after makeup removal in terms of image quality metrics: SSIM and PSNR, and reported our findings in Table~\ref{tab:visqual}. We compute the \textcolor{blue}{Original} performance by comparing the clean FFHQ face images (without makeup) with images simulated using EleGANt (with makeup). Our algorithm is at par with DiffAM in terms of SSIM=0.98 while being second best in terms of PSNR=35.64, showing high retention of visual fidelity. 

\begin{table}
\centering
\caption{Results of makeup removal in terms of image quality metrics (SSIM$\uparrow$ and PSNR$\uparrow$) on {FFHQ}. \textbf{Bolded} values indicate best results while \underline{underlined} indicates second-best results.}
\scalebox{0.75}{
\begin{tabular}{l||ll} \hline
& \textbf{SSIM} & \textbf{PSNR} \\ \hline \hline
\textcolor{blue}{Original}   &  0.58   &  14.94   \\
BeautyGAN  & 0.89    & 21.48    \\
LADN       & 0.43    & 13.53    \\
PSGAN++      & 0.85    & 20.65    \\
CLIP2Protect  & 0.46  & 15.74   \\
DiffAM      & \textbf{0.98}  & \textbf{39.52}  \\
MAD        &  0.92    &  26.46   \\ \hline
\begin{tabular}[c]{@{}l@{}}Ours-\textsc{DiffClean} (SSRNet)
\end{tabular} & \textbf{0.98}          & \underline{35.64}         \\
\begin{tabular}[c]{@{}l@{}}Ours-\textsc{DiffClean} (CLIP)
\end{tabular}    & \textbf{0.98}          & 35.29     \\ \hline   
\end{tabular}}
\label{tab:visqual}
\end{table}

\subsection{Real-world makeup generalizability}
Real-world use cases, such as online age verification \textit{only} include a facial image with or without makeup; it does not include the individual's actual age label. So, in this experiment, we applied \textsc{DiffClean} to real-world makeup images from the \textbf{BeautyFace}~\cite{Beautyface}, \textbf{LADN}~\cite{ladn_2019}, and \textbf{Makeup-Wild}~\cite{jiang2020psgan} datasets, and used MiVOLO to predict age before and after makeup removal. See examples of real-world makeup removal in Figs.~\ref{fig:realworld} and~\ref{fig:covariates}. Since we do not have ground truth age labels, instead, we report both overestimation and underestimation error statistics in Table~\ref{tab:realstats}. Further, we analyzed the confidence interval (CI) using a statistically significant sample size of 3K images from BeautyFace using the student's \textit{t}-distribution at a 95\% confidence level. We obtain CI = [2.88, 3.09] with a margin of error = 0.10 on the underestimation errors, and  CI = [2.75, 2.89] with a margin of error of 0.07 on the overestimation errors. Results indicate that \textsc{DiffClean} generalizes across various real-world makeup styles, pose, and illumination variations, and can be readily used in practical settings. 

\begin{table}
\centering
\caption{Statistics of predicted ages on the real-world makeup images from the {BeautyFace, LADN and Makeup-Wild} datasets before and after makeup removal, and associated overestimation and underestimation errors in predicted age.}
\scalebox{0.65}{
\begin{tabular}{l||llll} \hline
\textbf{\begin{tabular}[c]{@{}l@{}}Before makeup \\ removal\end{tabular}}          & \multicolumn{4}{c}{\textbf{After makeup removal}}                                                                                                                                                                \\ \cline{3-5}
$\mu \pm \sigma$                                                                           &                                                              & $\mu \pm \sigma$ & \begin{tabular}[c]{@{}l@{}}Overestimation\\ $\mu \pm \sigma$\end{tabular} & \begin{tabular}[c]{@{}l@{}}Underestimation\\ $\mu \pm \sigma$\end{tabular} \\ \hline \hline
\multirow{2}{*}{\begin{tabular}[c]{@{}l@{}}22.38 $\pm$ 3.36\\ {[}BeautyFace{]}\end{tabular}}   & \begin{tabular}[c]{@{}l@{}}\textsc{DiffClean}\\ (SSRNet)\end{tabular} &   21.28 $\pm$ 4.58       & 2.99 $\pm$ 2.92                                                                  &   2.79 $\pm$ 1.96                                                                 \\
                                                                                   & \begin{tabular}[c]{@{}l@{}}\textsc{DiffClean}\\ (CLIP)\end{tabular}   & 21.34 $\pm$ 4.66         & 3.14 $\pm$ 2.97                                                                  &   2.82 $\pm$ 1.95                                                                 \\ \hline
\multirow{2}{*}{\begin{tabular}[c]{@{}l@{}}24.64 $\pm$ 3.88\\ {[}LADN{]}\end{tabular}}         & \begin{tabular}[c]{@{}l@{}}\textsc{DiffClean}\\ (SSRNet)\end{tabular} & 22.59 $\pm$ 4.47         &   1.60 $\pm$ 1.11  &  3.02 $\pm$ 1.90   \\
    & \begin{tabular}[c]{@{}l@{}}\textsc{DiffClean}\\ (CLIP)\end{tabular}   &  22.55 $\pm$  4.51      & 1.62 $\pm$ 1.11                                                                  &  3.11 $\pm$ 1.97                                                                  \\ \hline
\multirow{2}{*}{\begin{tabular}[c]{@{}l@{}}25.56 $\pm$ 4.12\\ {[}Makeup-Wild{]}\end{tabular}} & \begin{tabular}[c]{@{}l@{}}\textsc{DiffClean}\\ (SSRNet)\end{tabular} & 24.16 $\pm$ 4.38    & 2.45 $\pm$ 2.31   &  3.34 $\pm$ 3.06  \\
    & \begin{tabular}[c]{@{}l@{}}\textsc{DiffClean}\\ (CLIP)\end{tabular}   & 24.25 $\pm$ 4.40  & 2.98 $\pm$ 2.80  & 3.20 $\pm$ 3.07 \\\hline                          

\end{tabular}}
\label{tab:realstats}
\end{table}

\subsection{Ablation study}

\textbf{Training Dataset Diversity.} 
We fine-tuned on 600 images (300 from MT dataset with smoky-eyes, flashy, retro, Korean and Japanese makeup styles + 300 from UTKFace with EleGANt-based makeup transfer with five reference styles and adaptive interpolated shading). To further assess training data diversity, we augmented our training set, $D_{tr}$, and report MAE on validation set, as follows: 4.42 ($\lvert D_{tr} \rvert=600$), 4.55 ($\lvert D_{tr} \rvert=1,200$) and 4.68 ($\lvert D_{tr} \rvert=1,600$). We observe consistent MAE across varying training data scale. \\
\noindent \textbf{Influence of each loss component.}
 We report MAE ($\downarrow$) on the validation set by fine-tuning separate models with loss components added incrementally, as follows: 9.28 ($\mathcal{L}_{clip}$), 7.43 ( $\mathcal{L}_{clip}+\mathcal{L}_{id}$), 4.70 ($\mathcal{L}_{clip}+\mathcal{L}_{id}+\mathcal{L}_{pips}$), and \textbf{4.42} ($\mathcal{L}_{clip}+\mathcal{L}_{id}+\mathcal{L}_{pips}+\mathcal{L}_{SSRNet-age}$). Evidently, the combined losses improve the overall performance.\\
\noindent \textbf{Demographic fairness analysis.}
Table~\ref{Tab:fair} reports demographic fairness analysis on FFHQ using DeepFace library~\cite{serengil2024deepface}, indicating \textsc{DiffClean} does not exhibit demographic disparity. FFHQ has more White images than Black and Indian images, but a balanced gender distribution.


\begin{figure}[h]
    \centering
    \includegraphics[width=0.7\columnwidth]{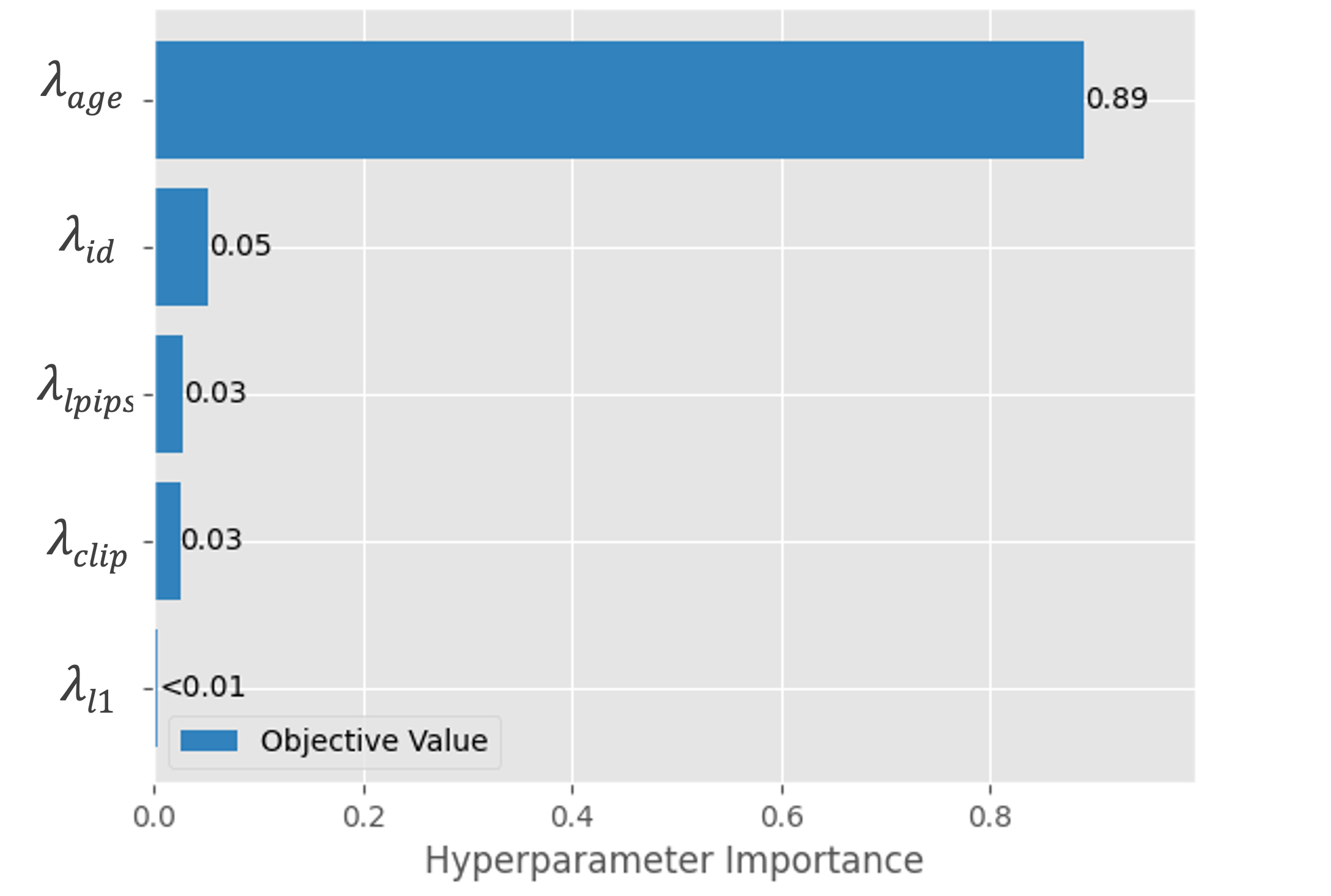}
    \caption{Importance of hyperparameters used in Eqn.~\ref{Eqn:totalloss}.}
    \label{fig:hyperparam}
\end{figure}

\begin{table}
\centering
\caption{Demographic fairness analysis on FFHQ using our \textsc{DiffClean} SSRNet/CLIP-age loss models in terms of MAE. Values in parentheses indicate count after being processed by DeepFace.}
\scalebox{0.74}{
\begin{tabular}{lllllll} \hline
\textbf{Race}   & \begin{tabular}[c]{@{}l@{}}White\\ (1,420)\end{tabular}    & \begin{tabular}[c]{@{}l@{}}Black\\ (23)\end{tabular}  & \begin{tabular}[c]{@{}l@{}}Asian\\ (507)\end{tabular}  & \begin{tabular}[c]{@{}l@{}}Hispanic\\ (254)\end{tabular} & \begin{tabular}[c]{@{}l@{}}Indian\\ (23)\end{tabular} & \begin{tabular}[c]{@{}l@{}}Middle\\ Eastern (129)\end{tabular} \\ \cline{2-7}
\textbf{MAE}    & 5.9/5.9 & 4.2/4.3 & 5.3/5.3 & 5.5/5.4  & 6.9/6.8 & 5.3/5.2                                                  \\ \hline \hline
\textbf{Gender} & \multicolumn{3}{c}{Male (1,188)}    & \multicolumn{3}{c}{Female (1,168)}  \\ \cline{2-7}
\textbf{MAE}    & \multicolumn{3}{c}{5.6/5.5} & \multicolumn{3}{c}{5.9/5.8} \\ \hline                     
\label{Tab:fair}
\end{tabular}}
\end{table}

\subsection{Practical Considerations and Further Analysis}
\label{subsec:AddAnalysis}
\begin{figure}[t]
    \centering
    \includegraphics[width=0.8\columnwidth]{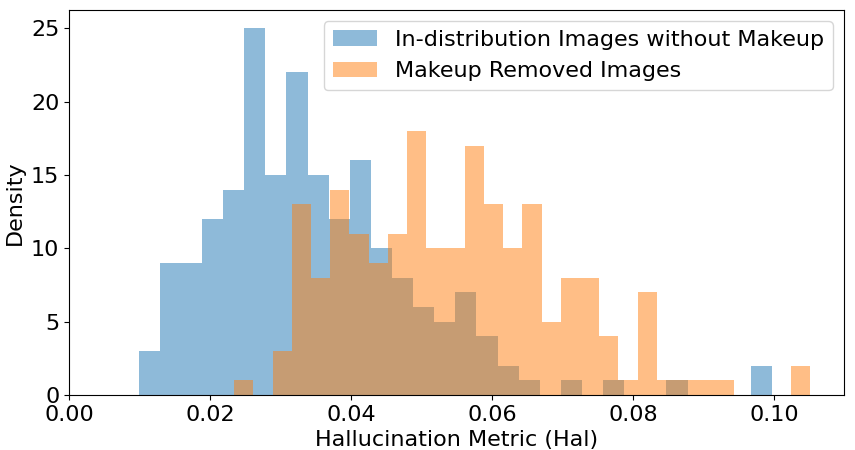}
    \caption{Histogram plot of hallucination metric~\cite{aithal2024understanding} for \textsc{DiffClean} generated images.}
    \label{fig:hal}
\end{figure}

 \noindent \textbf{Deployment scenario.} For real-world deployment, the model needs to accept an unconstrained facial image and remove makeup efficiently and effectively for facial analytics. \textsc{DiffClean} can handle pose, lighting variations, etc., and is faster (needs 2 secs.) than CLIP2Protect (30 secs.), MAD (3 mins.), and comparable with PSGAN++ (2.35 secs.) and DiffAM (2 secs.).\\
\noindent \textbf{Hyperparameter sensitivity.}
We used \textit{optuna}~\cite{akiba2019optuna}, an automated hyperparameter optimization framework, to perform multivariate sensitivity analysis to observe the interaction between the hyperparameters in Eqn.~\ref{Eqn:totalloss}. Hyperparameter importance plot after 30 trials shows that $\lambda_{age}$ has the highest importance=0.89 in the objective function; see Fig.~\ref{fig:hyperparam}. \\
\noindent \textbf{Effect of hallucinations.}
We computed the hallucination metric, $\text{Hal}(x)$, proposed in \cite{aithal2024understanding} to detect hallucinations in \textsc{DiffClean} generated makeup-removed images. We used 200 images from FFHQ test set over 80 timesteps, and computed: $\displaystyle \text{Hal}(x) = \frac{1}{|T_2 - T_1|} 
\sum_{i=T_1}^{T_2} 
 \left(\hat{x}_0^{(i)} - \overline{\hat{x}_0 ^{(t)}}\right)^2$
where $x_0^{(t)}$ represents the predicted values of the \textsc{DiffClean} outputs at $(t)$ time steps, and $\hat{x}_0 ^{(t)}$ is the mean predicted value from $T_1=0$ to $T_2=80$. Fig.~\ref{fig:hal} shows overlapping histogram plots of generated samples both for in-distribution data, \textit{i.e.}, no-makeup input (blue bars), and makeup-removed data, \textit{i.e.}, with-makeup input (orange bars), with mean values of \text{Hal}$(x)$ =0.04 and 0.06, respectively. 
Refer to Supplementary Materials for additional analysis.

\section{Conclusion}
In this work, we investigate the inadequacies of automated facial age estimators in presence of makeup. Makeup can be used to bypass age verification, exposing minors to cyber abuse. To address these issues, we designed \textsc{DiffClean}, a \textit{reference-free} diffusion-based makeup removal framework that can be deployed as a ready-to-plugin pre-processing module in a real-world application of online age verification for protecting minors. We combined text-guided CLIP loss with identity preservation and visual quality retention for makeup removal while restoring age-relevant features. Evaluations show \textsc{DiffClean} outperforms competing baselines in terms of age estimation (by 5.8\%) and identity verification (by 5.1\%) and is robust across real-world makeup styles.

{
    \small
    \bibliographystyle{ieeenat_fullname}
    \bibliography{main}
}

\maketitlesupplementary
\appendix
\section{Failure Case Analysis}

\begin{figure}[h]
        \centering
            \includegraphics[width=0.5\textwidth]{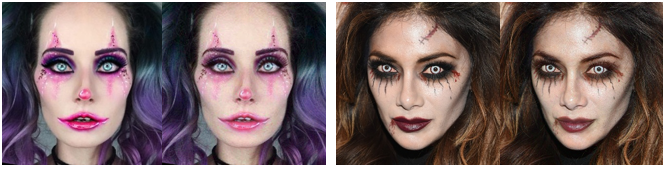} 
         \caption{Failure cases of \textsc{DiffClean} on extreme Halloween-style makeup images.}  
         \label{fig:failure}
\end{figure}

Fig.~\ref{fig:failure} shows examples of failure cases of makeup removal by \textsc{DiffClean} on the LADN dataset. Failure cases include drastic Halloween-style makeup, where our method can reduce the effect of makeup traces but not completely remove them. \textbf{In the future, we will use semantic segmentation masks to drive locally adaptive makeup removal against extreme makeup.}

\section{Additional evaluation} 
We present additional analysis by evaluating on new synthetic makeup data (IMDB-Clean), more results on baseline (MAD), and new face recognition model (IR-152).

\begin{figure*}[ht]
    \centering
    \includegraphics[width=0.95\textwidth]{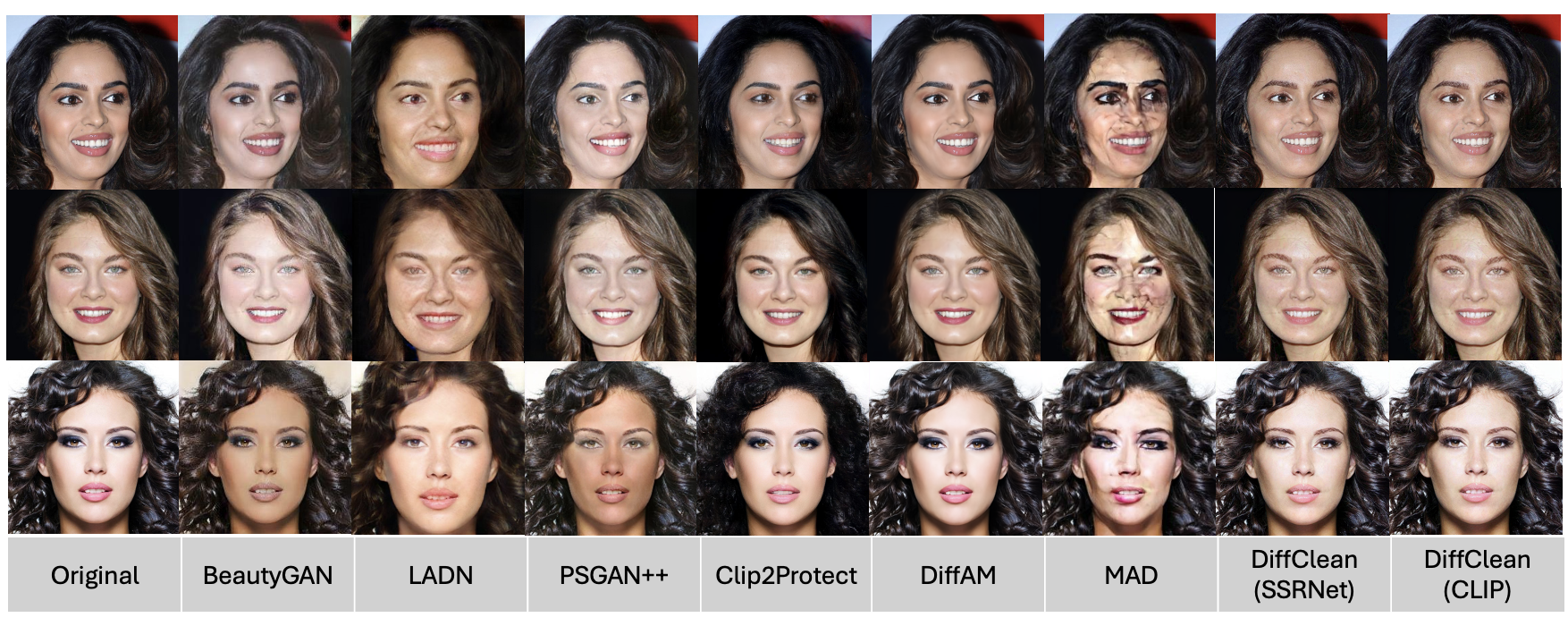}
    \caption{Comparison of makeup removal results generated by six baselines and our proposed \textsc{DiffClean} (last two columns) on three example images from CelebA-HQ \cite{karras2017progressive} dataset. GAN-based baselines (BeautyGAN, LADN, PSGAN++) introduce visual artifacts, while CLIP2Protect alters hair color and style, DiffAM does not effectively remove makeup, and MAD produces distortions on unseen data.}
    \label{fig:qual_old}
\end{figure*}

\textbf{(A) Data.} We test our method on a synthetic makeup dataset \textit{IMDB-Clean}~\cite{lin2022fp}, which is the age annotated version of IMDB-Clean. We used EleGANt~\cite{yang2022elegant} to curate the synthetic makeup transfer. We observe the following results. In terms of Minor/Adult accuracy($\uparrow$): 93.7\% (Makeup), 87.3\% (CLIP2Protect), \textbf{94.6}\% (Ours-SSRNet). In terms of  MAE($\downarrow$): 2.5 (Makeup), 3.4 (CLIP2Protect), \textbf{2.5} (Ours-SSRNet). Our method results in higher accuracy and lower MAE. We provide additional visual comparisons on example images from CelebA-HQ dataset in Fig.~\ref{fig:qual_old} 

\begin{figure}[h]
     \centering
    \includegraphics[width=0.45\textwidth]{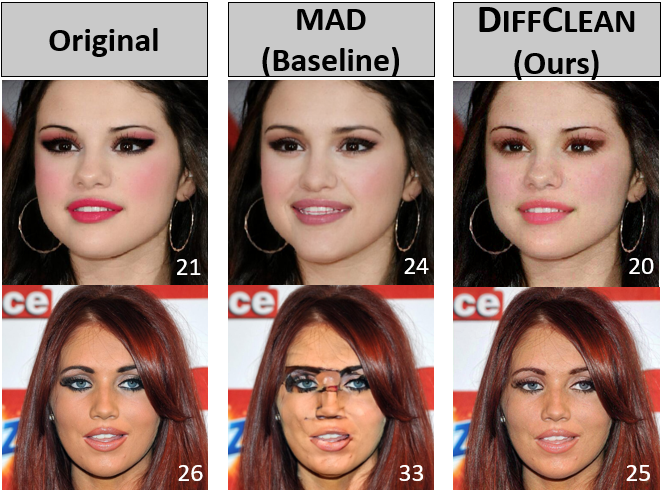}    
         \caption{Results of makeup removal on real-world makeup images from MT (top row) and CelebA-HQ (bottom row) datasets by MAD (baseline) and \textsc{DiffClean} (our method). Note our method is capable of restoring the correct age (indicated in white) compared to MAD.}  
         \label{fig:MAD}
\end{figure}

\textbf{(B) Baseline.} We present more examples of comparison between, \textsc{DiffClean} and MAD~\cite{ruan2025mad}; see Fig.~\ref{fig:MAD}. MAD struggles to restore original age and may introduce visible artifacts. The artifacts are more evident if the test dataset (CelebA-HQ) is different than the training dataset (MT) showing limited generalizability. In contrast, our method can handle cross-dataset and cross-style makeup removal. Even when tested on trained data (MT), it introduces hallucinations such as lower teeth in Fig.~\ref{fig:MAD} (1st row), which are not present in the original image.

\textbf{(C) Face recognition.} We computed the cosine similarity scores for the same identity (only genuine pairs) before makeup removal and after makeup removal using our method, \textsc{DiffClean} on the BeautyFace dataset using two different face matchers: IR-152 and MobileFace. We further compared it with the results of CLIP2Protect. The purpose of this experiment is to examine if our method introduces artifacts that inadvertently lowers the biometric similarity, resulting in false non-matches. As seen from the results in Table~\ref{Tab:facematcher}, our method produces higher genuine similarity compared to CLIP2Protect, thereby having lesser chances of false non-matches.

\begin{table}[]
\centering
\caption{Performance on BeautyFace dataset using IR-152 and MobileFace matchers in terms of genuine similarity scores($\uparrow$).}
\begin{tabular}{l||ll} \hline
\textbf{Method}                                                   & \textbf{IR-152} & \textbf{MobileFace} \\ \hline \hline
CLIP2Protect                                                      & 0.85            & 0.83                \\
\begin{tabular}[c]{@{}l@{}}DiffClean\\ (Ours-SSRNet)\end{tabular} & 0.92            & 0.95                \\
\begin{tabular}[c]{@{}l@{}}DiffClean\\ (Ours-CLIP)\end{tabular}   & 0.91            & 0.94   \\ \hline            
\end{tabular}
\label{Tab:facematcher}
\end{table}

\section{Real-world makeup data analysis}

\begin{figure}[h]
     \centering
     \begin{subfigure}[b]{0.4\textwidth}
         \centering
         \includegraphics[width=\textwidth]{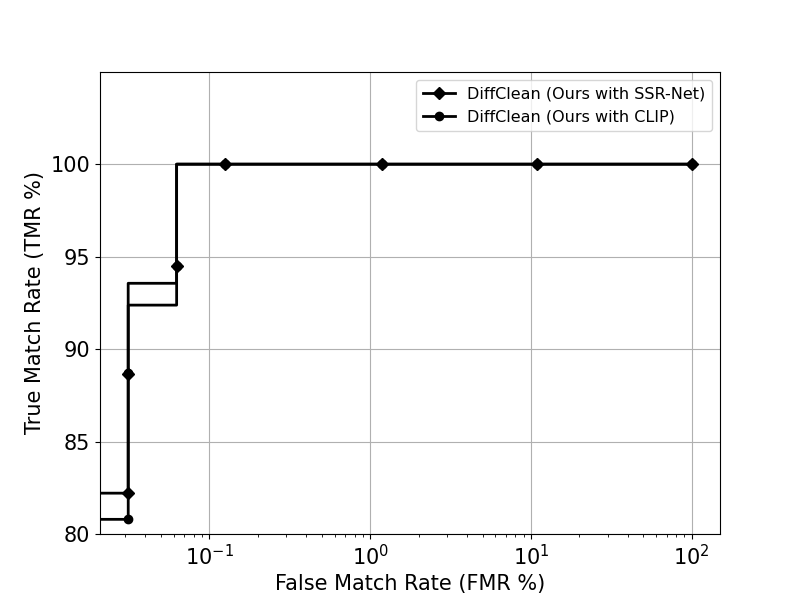} 
         \caption*{\textbf{FaceNet}}
     \end{subfigure} 
     \begin{subfigure}[b]{0.4\textwidth}
         \centering
         \includegraphics[width=\textwidth]{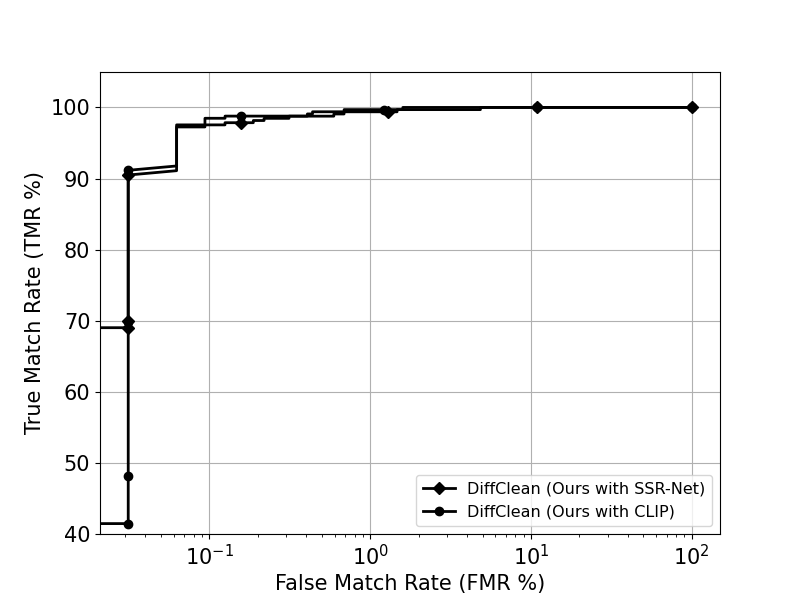}
         \caption*{\textbf{MobileFace}}
     \end{subfigure}  
    \begin{subfigure}[b]{0.48\textwidth}
    \centering
    \begin{tabular}{l||ll} \hline
     \multirow{2}{*}{\textbf{Method}} & \multicolumn{2}{c}{\textbf{FR model}}         \\ \cline{2-3}
        & FaceNet    & MobileFace  \\ \hline \hline
Ours-\textsc{DiffClean} (SSRNet)                 &   \textbf{81.6}       & \textbf{69.0}                 \\
Ours-\textsc{DiffClean} (CLIP)                   & {80.2}         & {41.5}                           \\ \hline 
\end{tabular}
    \end{subfigure}
  \caption{(Top): ROC curve with FaceNet. (Middle): ROC curve with MobileFace. (Bottom): Biometric matching in terms of TMR (\%) @FMR = 0.01\% (\textit{higher is better}) with FaceNet and MobileFace matchers on {LADN} dataset.}
    \label{fig:roc_ladn}
\end{figure}

We present the performance of our method in terms of identity verification on LADN dataset~\cite{ladn_2019} in Fig.~\ref{fig:roc_ladn} and image quality evaluation on LADN and Makeup-Wild datasets in Table~\ref{tab:img_qual_real}.


\begin{table}[]
\centering
\caption{Results of makeup removal in terms of image quality metrics (SSIM↑ and PSNR↑) on real-world makeup datasets LADN and Makeup-Wild.}
\label{tab:img_qual_real}
\resizebox{\columnwidth}{!}{%
\begin{tabular}{l||cc||cc}
\hline
\multirow{2}{*}{\textbf{Method}} & \multicolumn{2}{c||}{\textbf{LADN}} & \multicolumn{2}{c}{\textbf{Makeup-Wild}} \\ \cline{2-5}
& \textbf{SSIM} & \textbf{PSNR} & \textbf{SSIM} & \textbf{PSNR} \\ \hline\hline
Ours-\textsc{DiffClean} (SSRNet) & 0.94 & 30.77  & 0.97 & 35.09 \\
Ours-\textsc{DiffClean} (CLIP)   & 0.94 & 30.59 & 0.97 & 34.97          \\ \hline
\end{tabular}
}
\end{table}



\begin{table*}[]
\centering
\caption{Results of age estimation accuracy (\%) $\uparrow$ on each age group on the FFHQ dataset.}
\label{tab:ageest2}
\scalebox{0.85}{
\begin{tabular}{l||lllll} \hline
\textbf{Age group} & \textbf{Makeup images} & \textbf{\begin{tabular}[c]{@{}l@{}}CLIP2Protect\\  (‘no makeup’) prompt\end{tabular}} & \textbf{\begin{tabular}[c]{@{}l@{}}DiffAM \\  (MR)\end{tabular}} & \textbf{\begin{tabular}[c]{@{}l@{}}\textsc{DiffClean} (Ours)\\ SSRNet age loss\end{tabular}} & \textbf{\begin{tabular}[c]{@{}l@{}}\textsc{DiffClean} (Ours)\\ Clip age loss\end{tabular}} \\ \hline \hline
\textbf{0-2}   & 0.0  & 0.0  & 0.0  & 0.0  & 0.0  \\
\textbf{3-6}   & 2.1  & 1.8  & 2.8  & 1.8  & 1.4  \\
\textbf{7-9}   & 3.5  & 2.5  & 3.9  & 3.9  & 5.0  \\
\textbf{10-14} & 26.5 & 25.1 & 28.6 & 45.5 & 46.2 \\
\textbf{15-19} & 28.8 & 26.8 & 30.6 & 43.4 & 43.0 \\
\textbf{20-29} & 75.6 & 72.6 & 68.0 & 65.0 & 65.0 \\
\textbf{30-39} & 50.1 & 47.1 & 48.0 & 48.7 & 46.9 \\
\textbf{40-49} & 53.7 & 58.1 & 58.1 & 53.4 & 55.5 \\
\textbf{50-69} & 70.2 & 66.3 & 72.2 & 70.7 & 70.7 \\ \hline                                                                           
\end{tabular}}
\end{table*}

\section{Performance breakdown by age groups}
We present a detailed performance breakdown of \textsc{DiffClean} on each age group in Table~\ref{tab:ageest2}.
Note that CLIP-based age loss is marginally better than SSRNet-based age loss. Our method improves the age estimation accuracy on the \textit{target age} groups (minors/teenagers): [10-14] from 25\% (CLIP2Protect) and 28\% (DiffAM) to \textbf{46\%} (Ours); age group [15-19] from 26\% (CLIP2Protect) and 31\% (DiffAM) to \textbf{43\%} (Ours) as seen in Table~\ref{tab:ageest2}, while sacrificing accuracy in [20-29] age group by 8\%. On average, our method achieves the lowest mean and standard deviation of MAE across 9 age groups, as follows: $6.53 \pm 2.4$ (CLIP2Protect), $6.31 \pm 2.1$ (DiffAM), $5.75 \pm 1.7$ (Ours-SSRNet) and $\textbf{5.70} \pm \textbf{1.7}$ (Ours-CLIP).

We further investigated the drop in performance in the [20-29] group by comparing the ground-truth age with predicted age on original vs. makeup removed images in FFHQ. We observed MAE: 3.75 vs. 4.11, number of underestimation errors: 24 vs. 51,  number of overestimation errors: 64 vs. 45. This shows that our method lowers the overestimation errors in predicted age due to makeup at the expense of overall higher MAE.

\section{Importance of Weighted Self-Adjusted Smoothed L1 Loss Function}

Eqn.(1) in the main paper refers to the self-adjusted smoothed $\mathcal{L}_1$ loss function adopted in RetinaMask which is weighted by 3.0 for the vulnerable age groups between 10-29 yrs. It produced the lowest MAE compared to other losses; see \ref{Tab:difflosses}.

\begin{table}[]
\centering
\caption{Comparison of $\mathcal{L}_{WSL}$ with $\mathcal{L}_1$, $\mathcal{L}_2$, and Huber loss functions for age estimator.}
\scalebox{0.7}{
\begin{tabular}{cccccc} \hline
\multicolumn{6}{c}{\textit{MAE with different age losses}}                                                                             \\ \hline\hline
\textbf{Age Group} & \textbf{Pretrained} & \textbf{L1 Loss} & \textbf{MSE Loss} & \textbf{Huber Loss} & \textbf{$\mathcal{L}_{WSL}$ (Eqn.1)}         \\ \hline
\textbf{0-2}       & 20.2             & 2.1           & 3.1            & 2.6              & 1.6                         \\
\textbf{3-6}       & 27.9              & 2.3           & 2.6            & 2.5              & 2.2                         \\
\textbf{7-9}       & 29.2             & 3.7          & 4.1            & 3.7               & 3.1                         \\
\textbf{10-14}     & 26.4              & 5.0           & 5.1            & 5.1              & 4.6 \\
\textbf{15-19}     & 18.8             & 5.6            & 6.1            & 6.3               & 5.3 \\
\textbf{20-29}     & 14.6             & 6.4            & 6.2            & 6.5              & 6.1 \\
\textbf{30-39}     & 12.6             & 6.6           & 6.0            & 6.4              & 6.0                          \\
\textbf{40-49}     & 9.7              & 8.0           & 7.0            & 7.7              & 6.8                         \\
\textbf{50-69}     & 9.4              & 9.7           & 8.7            & 8.9              & 6.9                         \\ \hline
\textbf{Average}   & 18.7            & 5.5          & 5.4            & 5.5             & \textbf{4.7} \\ \hline
\end{tabular}}
\label{Tab:difflosses}
\end{table}

\begin{table}[t]
\centering
\caption{Attack Success Rate (ASR) ($\uparrow$) with MobileFace after makeup removal with \textsc{DiffClean} on protected faces generated using CLIP2Protect and DiffAM.}
\label{tab:ASR}
\begin{tabular}{lll}\hline
\multirow{2}{*}{\textbf{Methods}} & \multicolumn{2}{l}{\textbf{\begin{tabular}[c]{@{}c@{}}Attack Success Rate\\ (ASR)\end{tabular}}} \\
                        & CLIP2Protect          & DiffAM           \\ \hline\hline
Protected Faces               &  83.04   &   58.62                               \\ \hline
\textsc{DiffClean}               & \multirow{2}{*}{84.35 ($\uparrow$)}    & \multirow{2}{*}{61.45 ($\uparrow$) }   \\
(Ours-SSRNet)                       &                       &                       \\ \hline
\textsc{DiffClean}               & \multirow{2}{*}{84.56 ($\uparrow$)}   & \multirow{2}{*}{62.80 ($\uparrow$) }    \\ 
(Ours-CLIP)                       &                       &           \\\hline           
\end{tabular}%
\end{table}

\section{Impact on adversarial makeup transfer-based privacy protection}
Both CLIP2Protect and DiffAM are essentially makeup transfer-based privacy protection schemes that impersonate a targeted identity to fool the face matcher. \textsc{DiffClean} removes makeup using a diffusion model, and we wanted to investigate whether it affects the attack success rate (ASR)~\cite{sun2024diffam} of the protected faces on 
MobileFace (threshold=0.302) at False Match Rate=0.01 following~\cite{shamshad2023clip2protect}. $ASR=\frac{\# sim(protected face, target id)>th}{\# no. of comparisons}$. We conducted a preliminary analysis using 1,000 images from CelebA-HQ~\cite{karras2017progressive} dataset and one targeted identity for impersonation attack with 1 prompt for CLIP2Protect (`Matte') and 1 reference style from DiffAM (`XMY-060'). Our initial findings indicate that \textsc{DiffClean} is benign and does not compromise the privacy of protected faces (see Table~\ref{tab:ASR}). In the future, we will comprehensively analyze this impact on various protection mechanisms across multiple makeup styles.





\end{document}